\documentclass[conference]{IEEEtran}
\IEEEoverridecommandlockouts
\usepackage{cite}
\usepackage{amsmath,amssymb,amsfonts}
\usepackage{algorithmic}
\usepackage{graphicx}
\usepackage{textcomp}
\usepackage{xcolor}

\usepackage[utf8]{inputenc}   % für Umlaute
\usepackage{subcaption}

\def\BibTeX{{\rm B\kern-.05em{\sc i\kern-.025em b}\kern-.08em
    T\kern-.1667em\lower.7ex\hbox{E}\kern-.125emX}}
\begin{document}

\title{ShuffleNASNets: Efficient CNN models through modified Efficient Neural Architecture Search}
%\title{ShuffleNASNets: Efficient models through human knowledge and architecture search\\}

\author{
\IEEEauthorblockN{Kevin A. Laube}
\IEEEauthorblockA{\textit{Cognitive Systems Group} \\
\textit{University of Tübingen}\\
Tübingen, Germany \\
kevin.laube@uni-tuebingen.de}
\and
\IEEEauthorblockN{Andreas Zell}
\IEEEauthorblockA{\textit{Cognitive Systems Group} \\
\textit{University of Tübingen}\\
Tübingen, Germany \\
andreas.zell@uni-tuebingen.de}
}

\maketitle

\begin{abstract}
Neural network architectures found by sophistic search algorithms achieve strikingly good test performance, surpassing most human-crafted network models by significant margins.
Although computationally efficient, their design is often very complex, impairing execution speed. Additionally, finding models outside of the search space is not possible by design.
While our space is still limited, we implement undiscoverable expert knowledge into the economic search algorithm \textit{Efficient Neural Architecture Search (ENAS)}, guided by the design principles and architecture of \textit{ShuffleNet~V2}.
While maintaining baseline-like 2.85\% test error on CIFAR-10, our \textit{ShuffleNASNets} are significantly less complex, require fewer parameters, and are two times faster than the ENAS baseline in a classification task.
These models also scale well to a low parameter space, achieving less than 5\% test error with little regularization and only 236K parameters.
\end{abstract}

%\begin{IEEEkeywords}
%machine learning, image classification, neural networks
%\end{IEEEkeywords}

% ----------------------------------------------------------------

\section{Introduction}
\label{sec_introduction}

Image classification accuracies have been significantly improved in the past few years. After the success of AlexNet~\cite{intro_alex}, many faster and more accurate network structures have been proposed, such as VGG~\cite{intro_vgg}, GoogLeNet~\cite{intro_inc}, ResNet~\cite{o_res}, DenseNet~\cite{o_dense} and many more.
Yet designing successful neural network architectures often requires human experts to invest significant amounts of time and effort, in trial and error.
To reduce this investment, many ways of automating architecture search have been developed~\cite{s_par, m_darts, nas, nas_prog, nas_evo, nas_trans, enas, m_alphax, m_nash, m_nao, m_gcnn, m_smash}.

We focus on Efficient neural architecture search (ENAS)~\cite{enas}, a fast and highly economical algorithm from the Neural Architecture Search (NAS) family, which reduces the GPU hours from over 30000 to around 12.
In NAS, a recurrent controller network~\cite{nas, nas_trans, nas_prog, enas} samples candidate architectures, which are trained and evaluated. The resulting performance is used as a guiding signal to train the controller network, which will then find promising architectures more often. For efficiency, recent NAS algorithms only sample building blocks (cells), which are stacked to create a full architecture.

Despite their outstanding performance, measured by test accuracy, FLOPS, or parameter count, the high fragmentation of these network models has negative impacts on their inference time.
However, runtime is of importance for applications such as collision avoidance in autonomous driving or aerial vehicles control tasks.

In this work, we modify ENAS to adhere to several design guidelines that improve time performance~\cite{s_v2}. This is achieved by changing the design of sampled cells, as well as stacking the cells according to the design schema of ShuffleNet~V2~\cite{s_v2}.

We quickly and reliably find fast architectures that achieve less than 5\% error rate on CIFAR-10, often requiring less than one million parameters.
Our best found cell in this regard, making up configurations of ShuffleNASNet-A, requires a full model of only 236k parameters and little regularization for this achievement.
We find the cells for ShuffleNASNet-B to scale better with an increasing channel count, achieving up to 2.85\% test error, and thus being on par with the ENAS baseline.

%main contribution

% ----------------------------------------------------------------

\begin{figure*}[htbp]
	% left, bottom, right, top
	\begin{subfigure}{.35\textwidth}
	    \includegraphics[trim=31 910 318 25,clip,width=\textwidth]{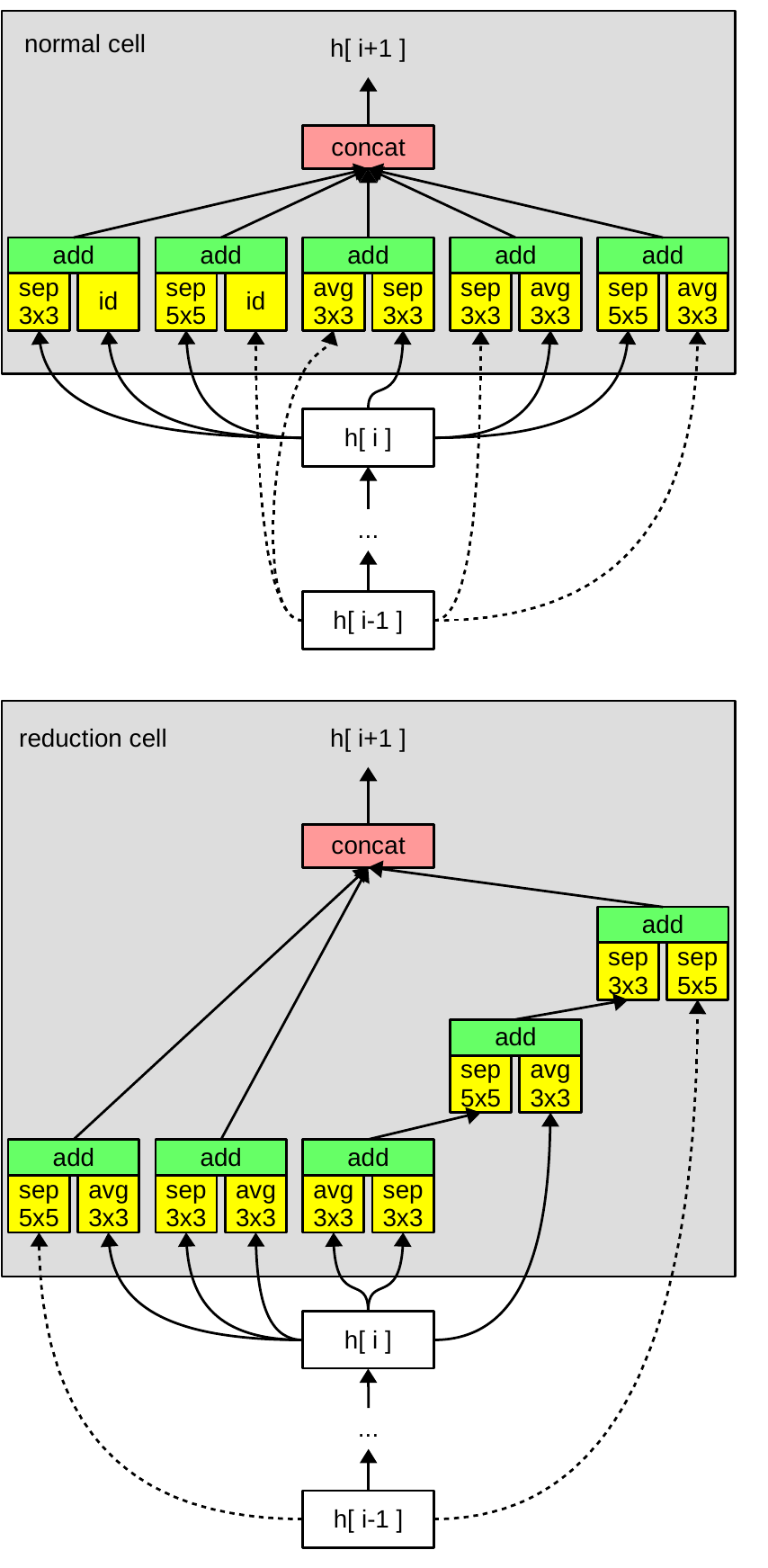}
	\end{subfigure}
	\hfill
	\begin{subfigure}{.63\textwidth}
		\includegraphics[trim=270 910 0 120,clip,width=\textwidth]{./img/cells.pdf}
	\end{subfigure}%
	%\centerline{\includegraphics[trim=0 321 0 25,clip,scale=0.85]{./img/cells.pdf}}
	\caption{Normal and reduction cells discovered by ENAS (left), ShuffleNASNet-A (middle), and ShuffleNASNet-B (right). The gray background marks what is considered a cell. We found ShuffleNASNet-B to achieve better results with the addition of batch normalization~\cite{etc_bn}. Bypassing 1x1* convolutions are factorized reductions and further explained in Section~\ref{secs_shortcut}.}
	\label{fig_cells}
\end{figure*}

\section{Related Work}
\label{sec_related_work}

\begin{figure*}[htbp]
	% left, bottom, right, top
	\centerline{\includegraphics[trim=36 490 90 855,clip,scale=1]{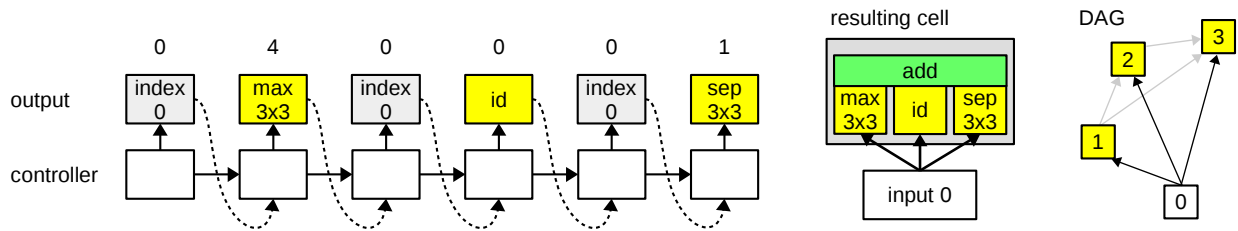}}
	\caption{The ENAS controller samples a series of $2B$ indices and $2B$ operation ids as integers, which are interpreted as building instructions for a cell. As  ShuffleNASNets cells are smaller, we only require $B$ indices and $B$ operations. The resulting cell is a subset of the DAG, connections that are not used are depicted in gray.}
	\label{fig_controller}
\end{figure*}

Our approach is directly related to ENAS~\cite{enas}, a fast and economical algorithm of the NAS family, of which automatically discovered architectures already achieved state-of-the-art on CIFAR-10, ImageNet and other benchmarks~\cite{nas_trans}~\cite{nas_evo}.
In ENAS, a recurrent controller of 100 LSTM units~\cite{etc_lstm} samples a series of integers, which are interpreted as building instructions for a CNN cell. $B$ blocks, sums of two operations each, are concatenated to form the cell output. Multiple cells with identical structure, but different weights and amount of filters, are stacked to create the network model.
For illustrations of model cells, sampling process and full architecture, see Figures~\ref{fig_cells},~\ref{fig_controller} and~\ref{fig_arc}, respectively.
The accuracy of the sampled model on a withheld validation set is used to guide the search algorithm to design better performing cells, using reinforcement learning~\cite{nas, enas, nas_trans, nas_prog} or evolution~\cite{nas_evo}.
While other NAS algorithms require thousands of models to be trained as guiding signal for their respective optimization method, ENAS considers a single directed and acyclic graph (DAG) that contains every possible cell.
Specific sampled cells share parameters in the DAG, which are reused when possible, significantly reducing the required training time in each iteration of the algorithm.\\

However, considering mobile applications, achieving good inference time is not as much in the focus~\cite{s_par} as accuracy and FLOPs, thus the currently most widespread models are usually handcrafted by human experts~\cite{s_v1, s_v2, s_mob2}.
Designed to achieve high accuracy even with limited computational power, memory and runtime, they often introduce clever operations and structures that are outside of the search spaces for automated discovery.
To improve time efficiency of our networks, we make use of the proposed shuffle operation and model structure of ShuffleNets~\cite{s_v2}~\cite{s_v1}, in which each cell receives only half of the channels as inputs, while the second half is skipped. The channels are shuffled after each layer, implicitly creating skip connections to later cells.
This design is simple yet powerful, and, similar to FractalNets~\cite{o_frac}, the model requires no identity functions to train.
We furthermore adhere to guidelines~\cite{s_v2} for time efficient networks, which we summarize:
\begin{itemize}
	\item G1) Equal channel width minimizes memory access cost
	\item G2) Excessive group convolution increases memory access cost
	\item G3) Network fragmentation reduces degree of parallelism
	\item G4) Element-wise operations are non-negligible
\end{itemize}

% ----------------------------------------------------------------

\section{Methods}
\label{sec_methods}

\subsection{Cell search space}

We employ ENAS to find promising candidate cells for our ShuffleNASNets, but make significant changes to the search space. Algorithms that operate in the \textit{NASNet search space}~\cite{nas_trans} sample a total of $2B$ operations and $2B$ input layer ids. These operations are applied to their corresponding input layers and summed pairwise in $B$ \textit{blocks}. Possible input layers are the two preceding cells as well as each block in the current cell that is already complete.
The published ENAS cells are shown in Figure \ref{fig_cells}.
\\
This design violates the guidelines mentioned in Section~\ref{sec_related_work}.
The massive fragmentation within and between cells increases the computation time, even though it is efficient considering FLOPs (G3).
Furthermore, operations acting on preceding cells typically have a factor of three to five times fewer outputs than input channels (G1).

We modify the search space to improve time efficiency for our models.
We use $B$ blocks (operations) with $B$ corresponding inputs, and use their sum as output. This simplification halves the fragmentation within cells (G3) and reduces the amount of element-wise operations considerably (G4).
We change operations to always use the same amount of input as output channels (G1), improving memory access costs. This also allows us to use identity functions that do not need 1x1 convolutions to correct for their mismatching channel count, except in reduction cells, where we require a stride of 2.
We add the 1x1 convolution as possible operation, so that normal cells can reorganize channels before they are shuffled.
For the current lack of an efficient group convolution implementation in TensorFlow~\cite{etc_tf}, we retain the depthwise separable operations of ENAS rather than following the ShuffleNet~V2 architecture.
This puts our search space to a total of six operations: we retain separable 3x3 and 5x5 convolutions as well as 3x3 max and min pooling, and add explicit identity and 1x1 convolution.
See Figure \ref{fig_cells} to easily compare resulting cells.
As in ENAS, separable 3x3 and 5x5 convolutions are applied twice and we make no use of bias parameters.

Furthermore, we modify which inputs the controller can pick from. While it is usual for NAS to receive two preceding cells as possible inputs, we only make half of the shuffled immediate prior layer available. This input implicitly contains skip connections to even more than two preceding cells. Naturally, operation results within the current cell are available as well. This reduces the fragmentation between cells (G3) and further simplifies the models, as spatial downsampling is now limited to reduction cells.

An interesting property of our search space is that for any $B_1 > B_2$, the space given by $B_1$ fully contains the space given by $B_2$. As the controller can simply use identity functions to pad existing paths within cells, there always exists an equivalent cell in all higher dimensional search spaces.
Other NAS methods may also produce equivalent cells, but are much less likely to discover them and will multiply one block result by a factor of two.
Additionally, in our design and unlike other NAS methods, the choice of $B$ has no influence on the number of output channels for each cell.

\begin{figure}[htbp]
	% left, bottom, right, top
	\centerline{\includegraphics[trim=25 620 340 600,clip,scale=1]{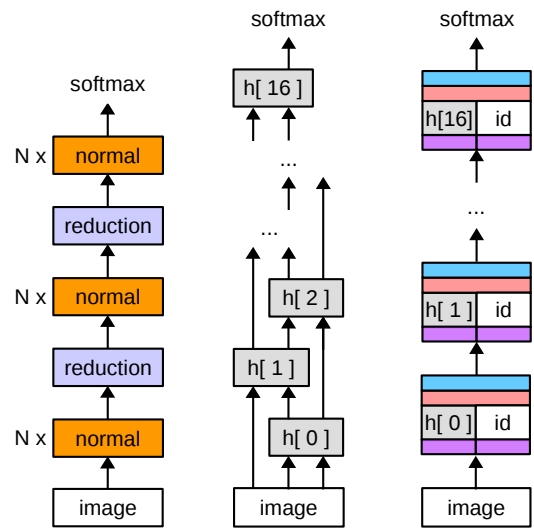}}
	\caption{Architecture of the overall CIFAR-10 models. Left: how NAS architectures are commonly presented; Middle: due to the cell design, NAS models are actually more complex; Right: ShuffleNASNet cells are embedded in split+concat+shuffle operations and therefore only use half of the available channels per cell.}
	\label{fig_arc}
\end{figure}

\subsection{Architecture}

Instead of simply stacking cells, as it is common for NAS~\cite{nas_trans, nas_prog, enas, nas_evo}, we make use of the efficient ShuffleNet~V2~\cite{s_v2} architecture and shuffle operation.
Both designs are illustrated in Figure~\ref{fig_arc}.

In each non-pooling layer, we split the input along the channel axis in two halves. The first half is used as cell input and then concatenated with the second, which remains unchanged. Finally, the channels are shuffled, so that another subset is used for the following cell.

Pooling layers simply use the full set of channels for both paths, doubling the amount whenever the spatial size is halved.
We explore the choice of how to reduce the spatial dimensions for bypassing channels in Section~\ref{secs_shortcut}, presenting three sensible approaches.

% ----------------------------------------------------------------

\begin{table*}[htbp]
	\caption{CIFAR-10 results with simple regularization (+, flipping, shifting, weight decay, drop-path) and additional Cutout (++). Models in the top group are handcrafted by human experts, models in the bottom group have been discovered automatically.}
	\begin{center}
		\begin{tabular}{ l | c c c | c  c }
			\hline
			\textbf{Method} & \textbf{\# layers} & & \textbf{\# params} & \textbf{test error+ (\%)} &  \\
			\hline
			
			FractalNet~\cite{o_frac} & 20 & & 38.6M & 4.60 &  \\
			PreAct ResNet~\cite{o_res} & 164 & & 1.7M & 5.46 &  \\
			DenseNet-BC~(k=12)~\cite{o_dense} & 100 & & 0.8M & 4.51 &  \\
			DenseNet-BC~(k=40)~\cite{o_dense} & 190 & & 25.6M & 3.46 &  \\
			WRN-40-4~\cite{o_wres} & 40 & & 8.9M & 4.53 & \\
			WRN-27-10~\cite{o_wres} & 28 & & 36.5M & 3.89 & \\
			PyramidNet~($\alpha$=48)~\cite{o_pyr} & & & 1.7M & 4.58 & \\
			PyramidNet~($\alpha$=84)~\cite{o_pyr} & & & 3.8M & 4.26 & \\
			\hline
			
			\multicolumn{4}{l}{} \\
			\hline
			\textbf{Method} & \textbf{\# cells} & \textbf{\# filters} & \textbf{\# params} & \textbf{test error+ (\%)} & \textbf{test error++ (\%)} \\
			\hline
			SMASHv1~\cite{m_smash} & & & 4.6M & 5.53 & \\
			DARTS (first order)~\cite{m_darts} & & & 2.9M & & 2.94 \\
			DARTS (second order)~\cite{m_darts} & & & 3.4M & & 2.83 \\
			DPP-Net-WS~\cite{s_par} & & & 1.0M & 4.78 & \\
			DPP-Net-M~\cite{s_par} & & & 0.45M & 5.84 & \\
			DPP-Net-Panacea~\cite{s_par} & & & 0.52M & 4.62 & \\
			\hline
			NASNet-A~\cite{nas_trans} & 18+2 & 32 & 3.3M & 3.41 & 2.65 \\
			PNAS-5~\cite{nas_prog} & 9+2 & 36 & 3.2M & 3.41 &  \\
			AmoebaNet-B~\cite{nas_evo} & 18+2 & 36 & 2.8M & 3.37 & 2.55 \\
			ENAS~\cite{enas} & 15+2 & 36 & 4.6M & \textbf{3.54} & \textbf{2.89} \\
			\hline
			ShuffleNASNet-A~(ours) & 12+2 & 36 & 0.24M & 4.93 & 4.31 \\
			ShuffleNASNet-A~(ours) & 15+2 & 48 & 0.47M & 4.40  & 3.85 \\
			ShuffleNASNet-A~(ours) & 15+2 & 64 & 0.80M  & 4.10  & 3.49 \\
			ShuffleNASNet-B~(ours) & 15+2 & 96 & 1.79M & 3.69 & 3.05 \\
			ShuffleNASNet-B~(ours) & 15+2 & 128 & 3.10M & \textbf{3.57} & \textbf{2.85} \\
			\hline
		\end{tabular}
		\label{tab_results}
	\end{center}
\end{table*}

\section{Experiments}
\label{sec_experiments}

\subsection{Experiment details}
\label{secs_details}

We evaluate ShuffleNASNets on the CIFAR-10~\cite{etc_cifar} dataset, which consists of 60000 32x32 pixel color images, each belonging to exactly one of 10 classes.
We hold out 10000 images as test set during search and final model training, and an additional 5000 images validation set for the search process.

We adopt standard data augmentation techniques that are widely used for this dataset. Training images are padded to 40x40, randomly cropped to 32x32, and randomly horizontally flipped.
Results with additional Cutout~\cite{etc_cutout} are reported separately and use a square of 16x16 pixels.
We use drop-path~\cite{o_frac} in all experiments and study its impact in Section~\ref{secs_merge}. 
All images are normalized by the mean and standard deviation of the dataset.

We use the training schedule of ENAS in all experiments, for search and final training.
During search, the controller parameters $\theta$ and the model parameters $\omega$ are optimized in turns. The controller is trained with Adam~\cite{etc_adam} and REINFORCE~\cite{etc_reinf}, based on the reward signal of the model. It is sufficient to sample and test ten cells on the validation set, with only one batch per cell.
Then the model, given one specific sampled cell, is trained for an entire epoch with stochastic gradient descent. The learning rates of both optimizers are subject to cosine annealing with warm restarts~\cite{etc_sgdr}, parametrized with $T_0=10$ and $T_{mult}=2$. We train each model for six annealing cycles, a total of 630 epochs, and use a batch size of 144 for search and final training.

\subsection{Results on CIFAR-10}
\label{secs_results}

We present the two most promising discovered cells in Figure~\ref{fig_cells} and use them to create the two ShuffleNASNet models A and B.
A summary of test errors of our ShuffleNASNets, recent architecture search methods and popular hand-crafted approaches can be found in Table~\ref{tab_results}.
We report the average test error of the last five epochs, averaged over three independent trials.
Despite the much simpler cell design and fewer parameters, ShuffleNASNet-B achieves equal test accuracy to the ENAS baseline.

Even though the NAS architectures are labeled to only use 36 (32) filters, their blocks are concatenated to a total of up to 180 (160).
We find that, depending on the discovered cells, the test error of our models saturates quicker and reaches maximum performance at around 128 filters.

\subsection{Time performance}
\label{secs_time}

We compare the speed of different ShuffleNASNet models in an inference task, and include the baseline ENAS model. Every configuration runs for 1000 consecutive forward passes, uses no data augmentation, and runs on a single Nvidia GTX 1080Ti GPU.

\begin{figure}[htbp]
	% left, bottom, right, top
	\centerline{\includegraphics[trim=3 -5 32 20,clip,scale=0.73]{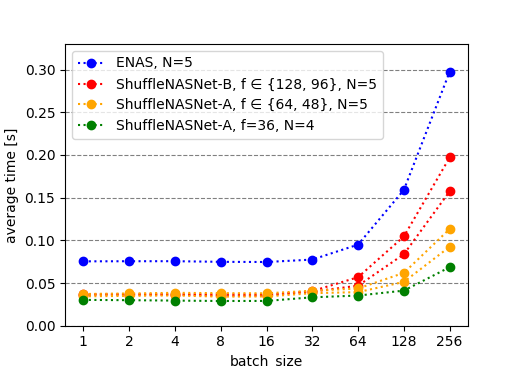}}
	\caption{Real-time requirements for forward passes of different models with varying batch size, number of filters $f$, and repetitions of normal cells $N$.}
	\label{fig_time}
\end{figure}

The results are displayed in Figure~\ref{fig_time}, highlighting a significant speed difference between ENAS and ShuffleNASNet architectures. Up to a batch size of 32, the number of channels in each model is of little importance.
Unsurprisingly, the main contributors to speed are a simple cell design and a low number of cells. 
Compared to the ENAS baseline, our ShuffleNASNet-B model is $2.05$ times faster. ShuffleNASNet-A models are similarly quick, $2.0$ times faster than ENAS with $N=5$, and $2.49$ times faster with $N=4$.

\subsection{Merge operation within cells and drop-path}
\label{secs_merge}

Contrary to a simple sum, we also experiment with concatenating the outputs of operations within a block, then applying a 1x1 convolution to correct the channel size. 
This not only violates guideline G1, it also significantly increases the parameter count of the resulting models. 

We searched for and tested several promising cells, and found that there is no advantage in test accuracy over using the simpler sum function.
The strongest performing cell discovered achieved $4.45\%$ test error with 64 channels (1.1M parameters) and $4.23\%$ test error with 96 channels (2.4M parameters), without using Cutout, performing worse than the simpler and similarly fast ShuffleNASNet-B models.

We experimented with drop-path and found the optimal drop rate varying significantly, depending on the applied merge operation.
Using a simple sum, unsurprisingly, keep probabilities of 0.9 to 1.0 work best for all tested cells. Contrary to that, adding batch normalization~\cite{etc_bn} causes the optimal keep probabilities to shift towards 0.5.
The stronger regularization effect is especially beneficial for models with more than one million parameters and vanishes for smaller ones, so that the additional batch normalization is not necessary for ShuffleNASNet-A models.

\begin{figure}[htbp]
	% left, bottom, right, top
	\centerline{\includegraphics[trim=56 330 280 1000,clip,width=0.5\textwidth]{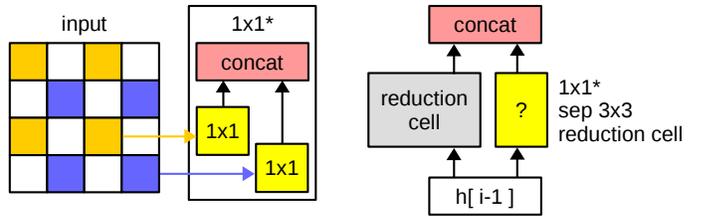}}
	\caption{Left: Factorized reduction concatenates two spatially shifted 1x1 convolutions. Right: possible shortcut operations.}
	\label{fig_shortcuts}
\end{figure}

\subsection{Bypassing reduction cells}
\label{secs_shortcut}

Tensors in reduction cells have their spatial dimensions reduced by a factor of two, making the use of an identity function for bypassing channels impossible.
We test three reasonable choices as shortcut operation:
\begin{itemize}
	\item A factorized reduction as used in ENAS, see Figure~\ref{fig_shortcuts}. Two spatially shifted 1x1 convolutions are concatenated, each convolution has half of the necessary channels as output and uses a stride of 2.
	\item Two stacked 3x3 depthwise separable convolutions, of which the first one uses a stride of 2. This is similar to the design of ShuffleNet V2.
	\item Using the reduction cell.
\end{itemize}

Using two reduction cell in each pooling layer increases the parameter count by around 17\%, resulting in roughly 16\% increased runtime.
Despite being the most powerful option, its accuracy is slightly worse across all tested cells. When implemented in ShuffleNASNet-B with 96 channels, the test error increases by 0.11\%, and we could not find similarly well performing networks when this shortcut operation was already part of the search process.
While the separable 3x3 convolution increases parameter count by around 3\% and runtime by only 2\%, similar as for a second reduction cell, the additional parameters decrease test performance by around 0.15\%.
Despite violating G1, we use the factorized 1x1 reduction for the superior parameter count, runtime, and test accuracy.

% ----------------------------------------------------------------

\section{Discussion}

% TODO more fluent smh
Contrary to ENAS, we find no advantage in using auxiliary heads during training. We speculate the model architecture and the low parameter count to be accountable for this, making an additional head of around 400K parameters unnecessary.

Compared to the ENAS baseline, our search space is significantly smaller. That may allow us to make changes in the future, such as adding possible operations, without the search space becoming too large to be efficiently traversed.
While the cell search is not fundamentally limited to the ENAS method, we chose it as starting point for its strong baseline and economic search properties.

Since larger search spaces (subject to $B$) contain all smaller spaces and keep the number of channels constant, the choice of $B$ can be used as an upper limit to the computational cost and parameter count of discovered models.

While using identity shortcuts is common practice since the outstanding success of ResNets~\cite{o_res}, we find it interesting that many well performing ShuffleNASNet cells implement more than one (see Figure~\ref{fig_cells}). 
One possible explanation is that, since our sum operation is not weighted, the search process compensates by using the identity operation twice.
We decided against weighted sums in accordance with G4, but may reconsider that decision in the future.

% TODO moar?
% Furthermore, the normal cell for the powerful ShuffleNASNet-B models is smaller than expected. While five paths are possible, it only uses four.

% ----------------------------------------------------------------

\section{Conclusion}

We have shown a modification of ENAS, based on ShuffleNets and their underlying design principles, resulting in models that are much simpler, faster, and require fewer parameters than the ENAS baseline. 
Nonetheless, our ShuffleNASNet-B models achieve equal test performance in the popular CIFAR-10 image classification task.
Also considering parameter efficiency, ShuffleNASNet-A models perform better than the strong DenseNet-BC~\cite{o_dense} and DPP-Net~\cite{s_par} baselines.

We achieve this by significantly changing the search space and network architecture, finding cells and creating models that other NAS algorithms can not discover.
We hope that in future work, further human knowledge is introduced to automated architecture search, simplifying the architectures and pushing the baselines of test accuracy, parameter efficiency and time performance.

%- evaluate on mobile platforms, e.g. drones
%- test e.g. with DoReFa net, extremely low bit versions with few params... high acc?

% ----------------------------------------------------------------


\begin{thebibliography}{00}

% intro alex
\bibitem{intro_alex} A. Krizhevsky, I. Sutskever and G. E. Hinton, ``ImageNet classification with deep convolutional neural networks'', In \textit{NIPS}, 2012.
% intro vgg
\bibitem{intro_vgg} K. Simonyan, A. Zisserman, ``Very deep convolutional networks for large-scale image recognition'', Arxiv, 1409.1556, 2014.
% intro inception
\bibitem{intro_inc} C. Szegedy, W. Liu, Y. Jia, P. Sermanet, S. Reed, D. Anguelov, D. Erhan, V. Vanhoucke and A. Rabinovich, ``Going deeper with convolutions'', In \textit{CVPR}, 2015.

% res net
\bibitem{o_res} K. He, X. Zhang, S. Ren, J. Sun, ``Identity Mappings in Deep Residual Networks'', Arxiv, 1603.050274, 2016.
% dense net
\bibitem{o_dense} G. Huang, Z. Liu, L. van der Maaten and K. Q. Weinberger, ``Densely Connected Convolutional Networks'', Arxiv, 1608.06993, 2016.


% nas survey
% \bibitem{nas_survey} T. Elsken, J. H. Metzen and F. Hutter, ``Neural Architecture Search: A Survey'', Arxiv, 1808.05377, 2018.
% nas
\bibitem{nas} B. Zoph and Q. V. Le, ``Neural Architecture Search with Reinforcement Learning'', Arxiv, 1611.01578, 2016.
% nas transferable
\bibitem{nas_trans} B. Zoph, V. Vasudevan, J. Shlens and Q. V. Le, ``Learning Transferable Architectures for Scalable Image Recognition'', Arxiv, 1707.07012, 2017.
% enas
\bibitem{enas} H. Pham, M. Y. Guan, B. Zoph, Q. V. Le and J. Dean, ``Efficient Neural Architecture Search via Parameter Sharing'', Arxiv, 1802.03268, 2018.
% nas prog
\bibitem{nas_prog} C. Liu, B. Zoph, M. Neumann, J. Shlens, W. Hua, L. Li, L. Fei-Fei, A. Yuille, J. Huang and K. Murphy, ``Progressive Neural Architecture Search'', Arxiv, 1712.00559, 2017.
% nas evo
\bibitem{nas_evo} E. Real, A. Aggarwal, Y. Huang and Q. V. Le, ``Regularized Evolution for Image Classifier Architecture Search'', Arxiv, 1802.01548, 2018.

% alphax
\bibitem{m_alphax} L. Wang, Y. Zhao and Y. Jinnai, ``AlphaX: eXploring Neural Architectures with Deep Neural Networks and Monte Carlo Tree Search'', Arxiv, 1805.07440, 2018.
% nash 
\bibitem{m_nash} T. Elsken, J. Metzen and F. Hutter, ``Simple And Efficient Architecture Search for Convolutional Neural Networks'', Arxiv, 1711.04528, 2017.
% nao
\bibitem{m_nao} R. Luo, F. Tian, T. Qin, E. Chen and T. Liu, ``Neural Architecture Optimization'', Arxiv, 1808.07233, 2018.
% genetic cnn
\bibitem{m_gcnn} L. Xie and A. Yuille, ``Genetic CNN'', Arxiv, 1703.01513, 2017.
% smash
\bibitem{m_smash} A. Brock, T. Lim, J.M. Ritchie and N. Weston, ``SMASH: One-Shot Model Architecture Search through HyperNetworks'', Arxiv, 1708.05344, 2017.
% darts
\bibitem{m_darts} H. Liu, K. Simonyan and Y. Yang, ``DARTS: Differentiable Architecture Search'', Arxiv, 1806.09055, 2018.
% dpp pareto
\bibitem{s_par} J. Dong, A. Cheng, D. Juan, W. Wei and M. Sun, ``DPP-Net: Device-aware Progressive Search for Pareto-optimal Neural Architectures'', Arxiv, 1806.08198, 2018.


% shufflenet v2
\bibitem{s_v2} N. Ma, X. Zhang, H. Zheng and J. Sun, ``ShuffleNet V2: Practical Guidelines for Efficient CNN Architecture Design'', Arxiv, 1807.11164, 2018.
% etc LSTM
\bibitem{etc_lstm} S. Hochreiter and J. Schmidhuber, ``Long short-term memory'', In \textit{Neural Computations}, 1997.
% shufflenet v1
\bibitem{s_v1} X. Zhang, X. Zhou, M. Lin and J. Sun, ``ShuffleNet: An Extremely Efficient Convolutional Neural Network for Mobile Devices'', Arxiv, 1707.01083, 2017.
% mobilenet v2
\bibitem{s_mob2} M. Sandler, A. Howard, M. Zhu, A. Zhmoginov and L. Chen, ``MobileNetV2: Inverted Residuals and Linear Bottlenecks'', Arxiv, 1801.04381, 2018.

% etc batchnorm
\bibitem{etc_bn} S. Ioffe and C. Szegedy, ``Batch Normalization: Accelerating Deep Network Training by Reducing Internal Covariate Shift'', Arxiv, 1502.03167, 2015.
% fractal net
\bibitem{o_frac} G. Larsson, M. Maire and G. Shakhnarovich, ``FractalNet: Ultra-Deep Neural Networks without Residuals'', Arxiv, 1605.07648v4, 2016.
% etc tensorflow
\bibitem{etc_tf} M. Abadi et al, ``TensorFlow: Large-Scale Machine Learning on Heterogeneous Systems'', https://www.tensorflow.org/
% etc cifar
\bibitem{etc_cifar} A. Krizhevsky and G. Hinton, ``Learning multiple layers of features from tiny images'', \textit{Tech Report}, 2009.
% etc cutout
\bibitem{etc_cutout} T. DeVries and G. W. Taylor, ``Improved Regularization of Convolutional Neural Networks with Cutout'', Arxiv, 1708.04552, 2017.
% etc ADAM
\bibitem{etc_adam} D. P. Kingma and J. L. Ba, ``Adam: A method for stochastic optimization'', In \textit{ICLR}, 2015.
% etc REINFORCE
\bibitem{etc_reinf} R. J. Williams, ``Simple statistical gradient-following algorithms for connectionist reinforcement learning'', \textit{Machine Learning}, 1992.
% etc cosine annealing warm restarts
\bibitem{etc_sgdr} I. Loshchilov and F. Hutter, ``SGDR: Stochastic Gradient Descent with Warm Restarts'', Arxiv, 1608.03983, 2016.
% wide res net
\bibitem{o_wres} S. Zagoruyko and N. Komodakis, ``Wide Residual Networks'', Arxiv, 1605.07146, 2016.
% pyramidal net
\bibitem{o_pyr} D. Han, J. Kim and J. Kim, ``Deep Pyramidal Residual Networks'', Arxiv, 1610.02915, 2016.



\end{thebibliography}
\end{document}